\documentclass[10pt,twocolumn,letterpaper]{article}

\usepackage[arxiv]{cvpr} 

\usepackage{amsmath,amsfonts}
\usepackage{algorithmic}
\usepackage{algorithm}
\usepackage{array}
\usepackage{textcomp}
\usepackage{stfloats}
\usepackage{url}
\usepackage{verbatim}
\usepackage{graphicx}
\usepackage{hyperref}
\usepackage{makecell}

\usepackage{algorithm}
\usepackage{algorithmic}

\usepackage{newfloat}
\usepackage{listings}
\usepackage{booktabs}
\usepackage{multirow}
\usepackage{graphicx}

\usepackage{natbib}

\title{Cross-View Image Set Geo-Localization}

\author{
Qiong Wu\textsuperscript{1}, Panwang Xia\textsuperscript{1}, Lei Yu\textsuperscript{2},  Yi Liu\textsuperscript{1}, Mingtao Xiong\textsuperscript{1},\\
Liheng Zhong\textsuperscript{2}, Jingdong Chen\textsuperscript{2}, Ming Yang\textsuperscript{2}\thanks{Corresponding authors.}, Yongjun Zhang\textsuperscript{1}, Yi Wan\textsuperscript{1}\footnotemark[1]~\\
\textsuperscript{1}Wuhan University \quad 
\textsuperscript{2}Ant Group \quad \\
{{\tt\small mabel\_wq@whu.edu.cn}}
}

\begin{document}
\maketitle
\begin{abstract}

Cross-view geo-localization (CVGL) has been widely applied in fields such as robotic navigation and augmented reality. Existing approaches primarily use single images or fixed-view image sequences as queries, which limits perspective diversity. In contrast, when humans determine their location visually, they typically move around to gather multiple perspectives. This behavior suggests that integrating diverse visual cues can improve geo-localization reliability. Therefore, we propose a novel task: Cross-View Image Set Geo-Localization (Set-CVGL), which gathers multiple images with diverse perspectives as \textbf{a query set} for localization. To support this task, we introduce SetVL-480K, a benchmark comprising 480,000 ground images captured worldwide and their corresponding satellite images, with each satellite image corresponds to an average of 40 ground images from varied perspectives and locations. Furthermore, we propose FlexGeo, a flexible method designed for Set-CVGL that can also adapt to single-image and image-sequence inputs. FlexGeo includes two key modules: the Similarity-guided Feature Fuser (SFF), which adaptively fuses image features without prior content dependency, and the Individual-level Attributes Learner (IAL), leveraging geo-attributes of each image for comprehensive scene perception. FlexGeo consistently outperforms existing methods on SetVL-480K and two public datasets, SeqGeo and KITTI-CVL, achieving a localization accuracy improvement of over 22\% on SetVL-480K. 
\end{abstract}    
\section{Introduction}
\label{sec:intro}

Cross-View Geo-Localization (CVGL) involves determining the geographical location of query data by matching it with images captured from different viewpoints, serving as an auxiliary tool for localization when GPS signals are not available \cite{wilson2024image}. It has broad applications in real-world scenarios, including robotic navigation \cite{zhu2021deep,boroujeni2024comprehensive,kabir2025terrain}, autonomous driving \cite{hane20173d,kim2017satellite}, and augmented reality \cite{chiu2018augmented}. 

The main challenge of CVGL is the substantial scale and appearance differences between images caused by viewpoint variations. Over the past few decades, researchers have made significant progress in CVGL by introducing advanced image encoders and optimization strategies~\cite{lin2022joint}. However, most of these efforts have only focused on single-image-based CVGL \cite{lin2013cross,shi2020looking}, without employing multiple query data. Other image-sequence-based studies~\cite{vyas2022gama,zhang2023cross,shi2022cvlnet} considered additional contents by introducing image sequences as queries, but the perspective diversity is still limited due to the typically fixed camera view in video sequences. Furthermore, these methods commonly rely on sequential relationships, which may not be available, as users cannot always provide images in a sequential order. 

\begin{figure}[t]
    \centering
    \includegraphics[width=1\linewidth]{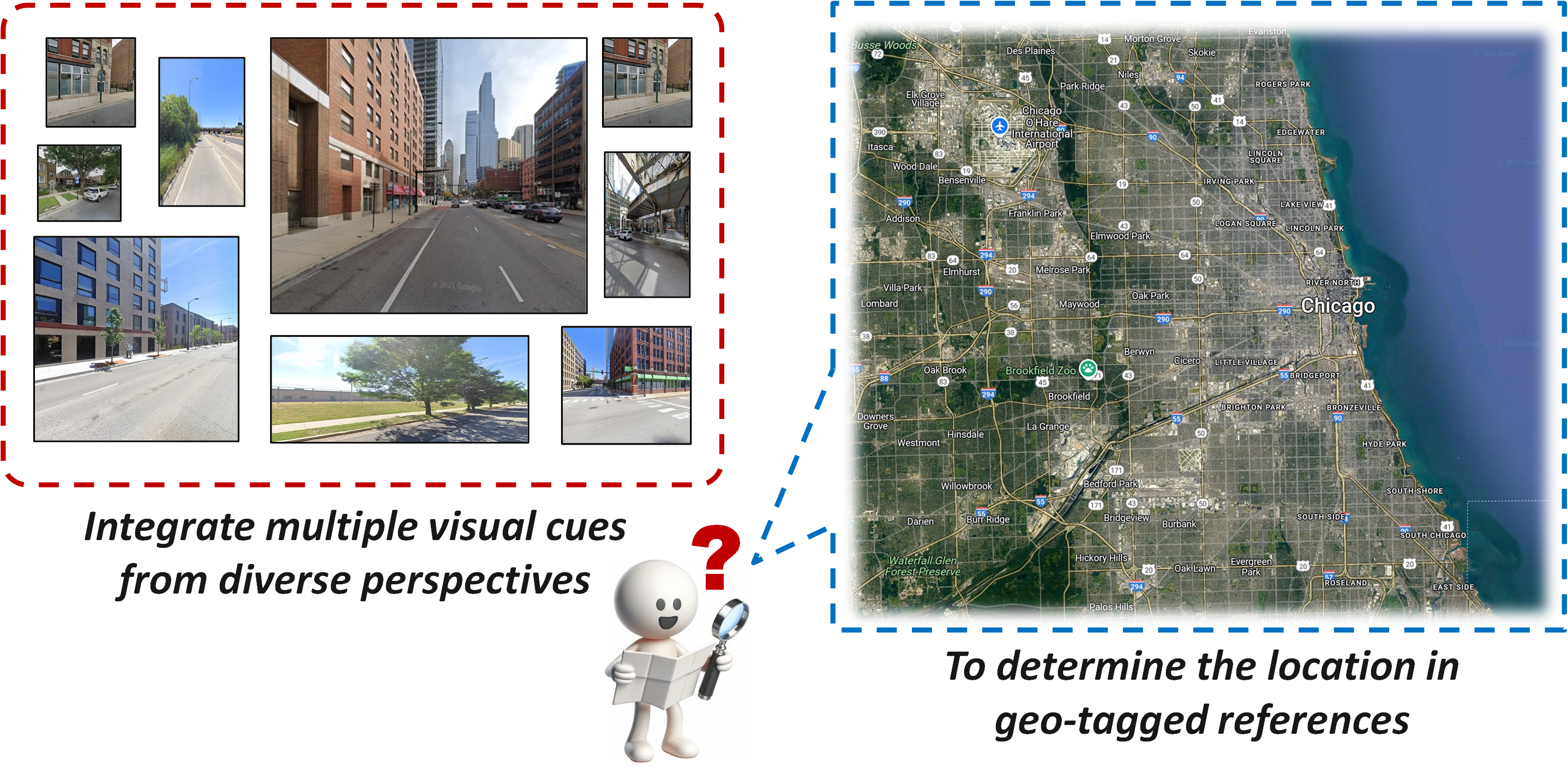}
    \vspace{-1.7em}
    \caption{Illustration of the Set-CVGL Task: Using multiple ground images captured from diverse perspectives to match geo-tagged references for location determination. }
    \label{fig_task_define}
   \vspace{-1.4em}
\end{figure}

When people determine their location through observation, they instinctively look around or even walk a short distance to gather more understanding of their surroundings. This natural behavior indicates that integrating multiple visual cues can enhance localization accuracy. Inspired by this observation, we propose a new task: Cross-View Image Set Geo-Localization (Set-CVGL), which uses images from multiple perspectives as a query set without sequential order and determines the location from geographic references, as illustrated in \cref{fig_task_define}. To support this new task, we introduce SetVL-480K, a dataset containing 480,000 ground images from six cities across different continents, with corresponding satellite images covering these areas. Unlike previous datasets based on video sequences, SetVL-480K's ground images are independently captured rather than by in-vehicle fixed cameras, offering diverse perspectives and sampling locations. On average, each satellite image cell aligns with 40 ground images, significantly more than other datasets (typically fewer than 10). By incorporating dense, unordered, and limited FOV images, SetVL-480K provides a realistic testbed for developing and evaluating methods capable of handling image sets.

Furthermore, we propose FlexGeo, a flexible method designed for Set-CVGL that can also adapt to single-image and image-sequence inputs. FlexGeo incorporates two key modules: the Similarity-guided Feature Fuser (SFF) and the Individual-level Attributes Learner (IAL). SFF fuses multiple features without relying on sequential relationships, adapting the weights of features to preserve distinct characteristics. IAL leverages geo-attributes of images for comprehensive scene perception, alleviating the information loss through multi-task learning.

Our main contributions can be summarized as follows: 

\begin{itemize}
\item We propose a novel task, Cross-View Image Set Geo-Localization (Set-CVGL), which gathers multiple images with diverse perspectives as a query set for cross-view geo-localization. To support Set-CVGL, we introduce a large-scale dataset, SetVL-480K, comprising 480,000 ground images and corresponding satellite images from six cities worldwide.
\item We propose FlexGeo, a method designed for Set-CVGL that is also adaptable to sequential or single images. It features two novel modules: the Similarity-guided Feature Fuser (SFF) and Individual-level Attributes Learner (IAL). SFF adaptively fuses features without relying on sequential relationships, while IAL leverages geo-attributes in each image to enhance scene understanding.
\item Our method outperforms existing methods on SetVL-480K, achieving over a 22\% improvement in localization accuracy. FlexGeo also achieves state-of-the-art performance on two public image-sequence-based datasets: SeqGeo and KITTI-CVL. Additionally, our experiments demonstrate that using multiple query images with diverse perspectives significantly boosts localization accuracy.
\end{itemize}

\section{Related Work}
\label{sec:related_work}
Cross-view geo-localization has gained widespread attention recently for its promising applications. We briefly review related works, categorizing them into single-image-based (S-CVGL) and image-sequence-based (Sequence-CVGL).

\subsection{Single-Image-Based}

As a core task in CVGL, single-image-based studies have garnered significant attention from scholars. Early studies \cite{lin2015learning,tian2017cross} addressed ground-to-aerial localization with paired image datasets. Building on this, large-scale datasets like CVUSA \cite{workman2015wide} and CVACT \cite{liu2019lending} were developed, using ground-view panoramas as queries and satellite images as references. Zhu \etal\cite{zhu2021vigor} introduced the VIGOR dataset, enabling seamless sampling of reference images. Zheng \etal\cite{zheng2020university} proposed University-1652 to study drone-view localization, and they then expanded it in varied environments \cite{wang2024multiple}. Dai \etal\cite{dai2023vision} introduced DenseUAV for self-positioning tasks using dense down-looking camera sampling. Fervers \etal\cite{fervers2025statewide} constructed a large-scale dataset to investigate scale variations in geo-localization.

Beyond dataset development, numerous methods have been advanced to tackle S-CVGL challenges. Early approaches \cite{castaldo2015semantic,bansal2011geo,lin2013cross} used handcrafted operators for cross-view feature extraction. With the advent of deep learning, Workman \etal\cite{workman2015location} pioneered using pre-trained AlexNet~\cite{krizhevsky2012imagenet} for scene understanding, later fine-tuning it on paired images \cite{workman2015wide}. This approach inspired a series of works focused on enhancing the discriminative capability of feature extractors through fine-tuning. 
To reduce the visual gap from viewpoint differences, some works applied transformations or alignments~\cite{liu2019lending,hu2018cvm}, such as optimal transport~\cite{shi2020optimal}, polar transformation~\cite{shi2019spatial}, region-level alignment~\cite{dai2021transformer}
and latent correspondences alignment~\cite{xia2024dac}. Meanwhile, other studies emphasized distinguishing features from different scenes \cite{lin2022joint,wang2022learning}. Shen \etal\cite{shen2023mccg} utilized cross-dimension interactions to improve feature differentiation, while Deuser \etal\cite{deuser2023sample4geo} used hard-negative sampling to strengthen scene distinction. Wu \etal\cite{wu2024camp} further explored intra-platform negative samples to enhance contrast. Additionally, unsupervised S-CVGL has gained interest \cite{li2024learning}. Li \etal\cite{li2024unleashing} investigated the utilization of unlabeled data in unsupervised settings.
These works have significantly advanced S-CVGL. However, they only focused on using single images as queries, neglecting the potential of multiple query images in geo-localization.

\subsection{Image-Sequence-Based}

Recently, several studies have recognized the value of using multiple images for queries, leading to the development of datasets and methods based on video sequences. Vyas \etal\cite{vyas2022gama} extended the BDD100k \cite{yu2020bdd100k} by incorporating aerial images to create the GAMa dataset for Sequence-CVGL. They introduced GAMa-Net along with a hierarchical approach for localizing long videos and short clips. 
Shi~\etal\cite{shi2022cvlnet} developed KITTI-CVL by augmenting the KITTI~\cite{geiger2013vision} dataset with satellite images. They proposed CVLNet, a network featuring a geometry-driven view projection module and a photo-consistency constrained sequence fusion module for handling image sequences. Zhang \etal\cite{zhang2023cross} introduced SeqGeo, a large-scale Sequence-CVGL dataset consisting of frontal camera images extracted from road vehicles and corresponding satellite images. They proposed an end-to-end temporal feature aggregation method for Sequence-CVGL. Pillai \etal\cite{pillai2024garet} proposed a fully transformed-based CVGL method to efficiently aggregate image-level representations and adapt it for video inputs.

Although these works have advanced CVGL, ground images in these datasets exhibit high content redundancy, and the methods are heavily reliant on sequential relationships. Consequently, unordered image sets in CVGL remain underexplored. To bridge this gap, this paper introduces Set-CVGL, along with a new benchmark and a method to support the task.

\section{Cross-View Image Set Geo-Localization}

This section defines the Set-CVGL task and then introduces the SetVL-480K dataset.

\begin{table*}[t]
    
            \vspace{-0.5em}
  
    \centering
    \resizebox{\textwidth}{!}{
    \begin{tabular}{l|ccccc} 
    \toprule
       \textbf{Dataset Comparison}   & VIGOR~\cite{zhu2021vigor}&GAMa~\cite{vyas2022gama}&  SeqGeo~\cite{zhang2023cross}&  KITTI-CVL~\cite{shi2022cvlnet} & SetVL-480K(Ours)\\ 
          \midrule
          Number of Query Images& 105,214& \(\sim \)15,384,000&  118,549&  28,740& 480,000\\ 
          Number of Reference Images& 90,618&1,923,000&  38,863&  41,463$^{\dagger}$& 16,530\\
 Queries per Reference&  \(\sim \)1& \(\sim \)8&  \(\sim \)7& 1& \(\sim \)40\\
 Geo-location Distribution& US& US& US& Karlsruhe,Germany&Worldwide\\
 Perspective Diversity& Panoramic& Limited& Limited& Limited&Free\\ 
 Limited FOV& \(\times \)&\(\surd \) & \(\surd \) & \(\surd \) &\(\surd \)\\
 Sequential Relationships&  \(\times \)&\(\surd \)& \(\surd \)& \(\surd \)& \(\times \)\\
   \bottomrule
    \end{tabular}}
    \caption{Comparison between the proposed SetVL-480K dataset and existing datasets. $^{\dagger}$ indicates that there are duplicate reference images, so the actual number of unique reference images is lower than stated.}
      \label{tab:dataset comparison}
\vspace{-1.0em}
\end{table*}

\subsection{Task Definition}

The Set-CVGL task extends the traditional CVGL problem to accommodate multiple query images from diverse perspectives without relying on sequential priors.

Given a set of query images captured within a certain radius from an unknown location and a database of geo-tagged reference images, the goal is to identify the reference image that best matches the query set, thereby determining its location.
Let \(M\) represent the total number of scenes. For a specific scene \(i\), the set of \(n\) ground images is denoted as \(X_{i}^{(n)} = \left \{ x_{i_1}, x_{i_2}, ..., x_{i_n} \right \} \), where \(\ n\ge1 \ \), and \(max(d(x_{i_{p}},x_{i_{q}}))\le R\) for all \(x_{i_{p}},x_{i_{q}}\in X_{i}^{(n)} \), indicating that all images are taken within radius \(R\). The reference database \(Y^M=\left \{ y_1, y_2, ..., y_M \right \} \) contains satellite images for each scene.

The objective of the Set-CVGL task is to identify the reference image \(y_{i}\) that best matches the set of query images \(X_{i}^{(n)} \), based on their feature representations in feature space with a similarity metric \(sim(X, y)\). Formally, this can be expressed as \cref{eq:problem_formulation1}:
\begin{equation}
y_i = \underset{j \in \{1, \ldots, M\}}{\arg\max} \, sim( F_{x}(X_{i}^{(n)}), F_{y}(y_j))
\label{eq:problem_formulation1}
\end{equation}
Here, \(F_{x}\left ( \cdot \right ) \) and \(F_{y}\left ( \cdot \right ) \) are the feature extraction functions for query image sets and reference images, respectively.

Set-CVGL enables a comprehensive capture of environmental content, mirroring human natural behavior. Meanwhile, it presents challenges, particularly in fusing features from multiple perspectives without sequential priors or high overlap.

\subsection{SetVL-480K Dataset}

\begin{figure}[t]
    \centering
    \includegraphics[width=1\linewidth]{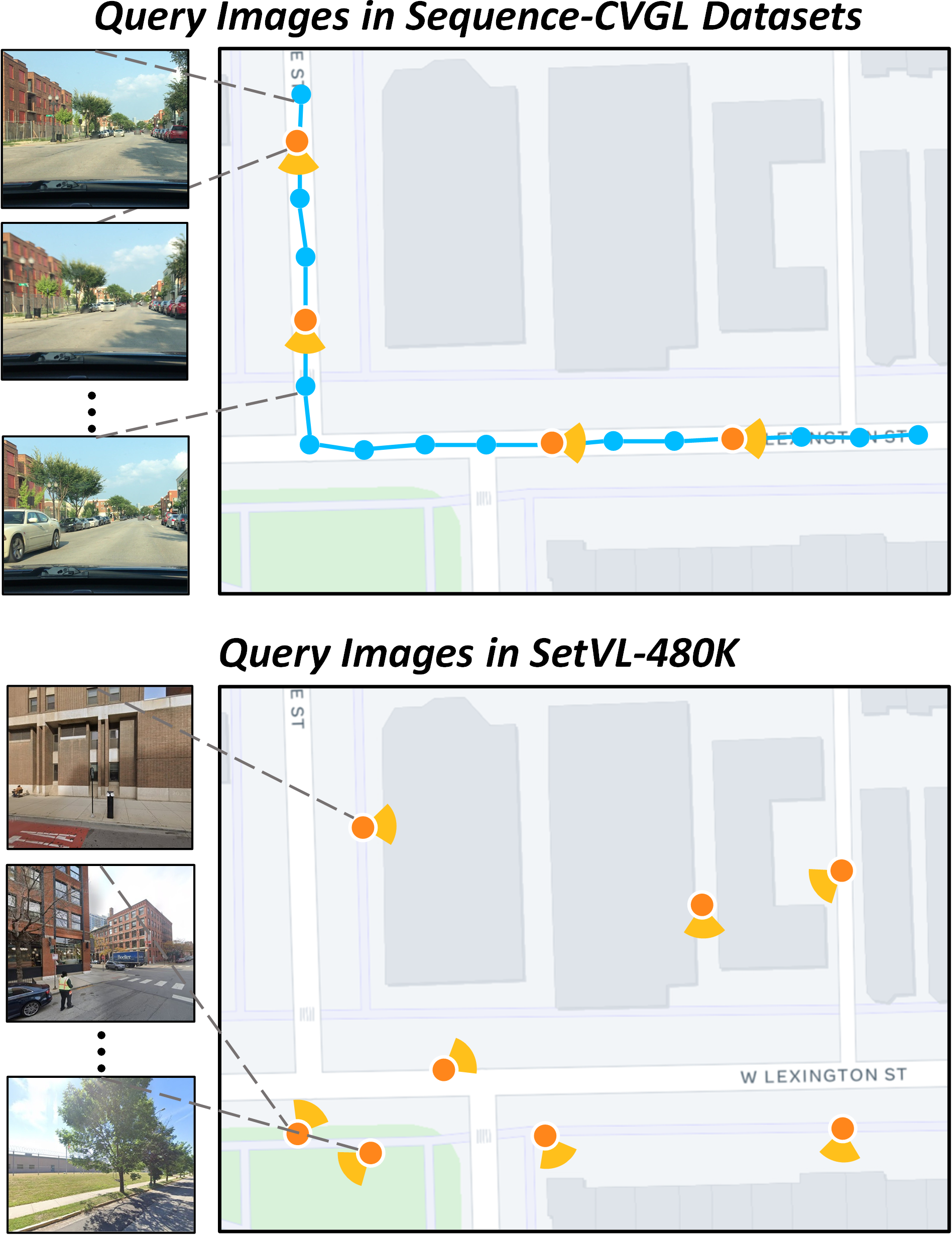}
    \vspace{-1.0em}
    \caption{Distribution of query data across different datasets. The top part illustrates query image distribution in Sequence-CVGL datasets, while the bottom part shows the distribution in SetVL-480K. The blue line represents the vehicle's route, with blue dots marking sampling points and orange markers indicating the direction of image capture.}
    \label{fig_sampling_comparison}
    \vspace{-2.0em}
\end{figure}

\noindent\textbf{Dataset Characteristics.} The primary aim of the SetVL-480K dataset is to support the Set-CVGL, a task we propose that remains underexplored in previous research. \cref{tab:dataset comparison} compares SetVL-480K with several previous datasets, highlighting its key characteristics:
\begin{enumerate}
    \item Diverse perspectives and sampling locations: As shown in \cref{fig_sampling_comparison}, ground images in Sequence-CVGL datasets often have highly repetitive views and limited sampling locations due to the road-bound trajectories and fixed camera view, leading to redundancy. In contrast, images in SetVL-480K are taken from varied perspectives and locations.
    \item  No sequential constraints among query images: Ground images in SetVL-480K are captured from random perspectives, presenting a general and challenging geo-localization task suited to a wide range of applications.
    \item Dense distribution of query images: In SetVL-480K, each reference cell is linked to an average of 40 query images, allowing researchers to investigate the impact of query image quantity on localization accuracy.

\end{enumerate}

\noindent\textbf{Dataset Construction.} The SetVL-480K dataset consists of 16,530 satellite images and 480,000 ground images, covering the central areas of six cities across the world: Chicago, Johannesburg, London, Rio de Janeiro, Sydney, and Taipei.

The satellite images, sourced from Esri World Imagery~\cite{aerialview}, have a \(400\times400\) resolution at zoom level 18, providing a ground resolution of approximately 0.597 meters per pixel. These images are cropped with an overlap of 0.125 for seamless coverage. Ground images, obtained using the Google Street View Static API \cite{streetview}, have a resolution of \(512\times512\) pixels, with a limited FOV of \(90^\circ\) rather than being panoramic. Following previous studies TouchDown~\cite{chen2019touchdown} and LHRS-Bot \cite{muhtar2024lhrs}, we provide all the ground image IDs and the processing scripts rather than the ground images themselves\footnote{\url{https://research.google/blog/enhancing-the-research-communitys-access-to-street-view-panoramas-for-language-grounding-tasks/}}. The training and testing sets are evenly divided, with a 1:1 ratio.

\noindent\textbf{The Evaluation Protocol.} For Set-CVGL, we adopt the evaluation metrics Recall@K-N and AP-N, which are the adaptations of the standard Recall@K and AP (Average Precision) metrics. The parameter N indicates the number of query images in a set. Recall@K measures the proportion of queries with the correct result in the top K retrieved results, reflecting localization accuracy within a rank range. Recall@K-N extends this by evaluating performance when aggregating N query images, capturing both retrieval rank K and query image count N. Similarly, AP-N assesses retrieval accuracy with multiple query images.

\section{Methodology}

In this section, we introduce FlexGeo, our method for Set-CVGL. We begin with an overview of FlexGeo’s framework, then detail two key modules: the Similarity-guided Feature Fuser (SFF) and the Individual-level Attributes Learner (IAL).

\begin{figure*}[t]
    \centering
    \includegraphics[width=1\linewidth]{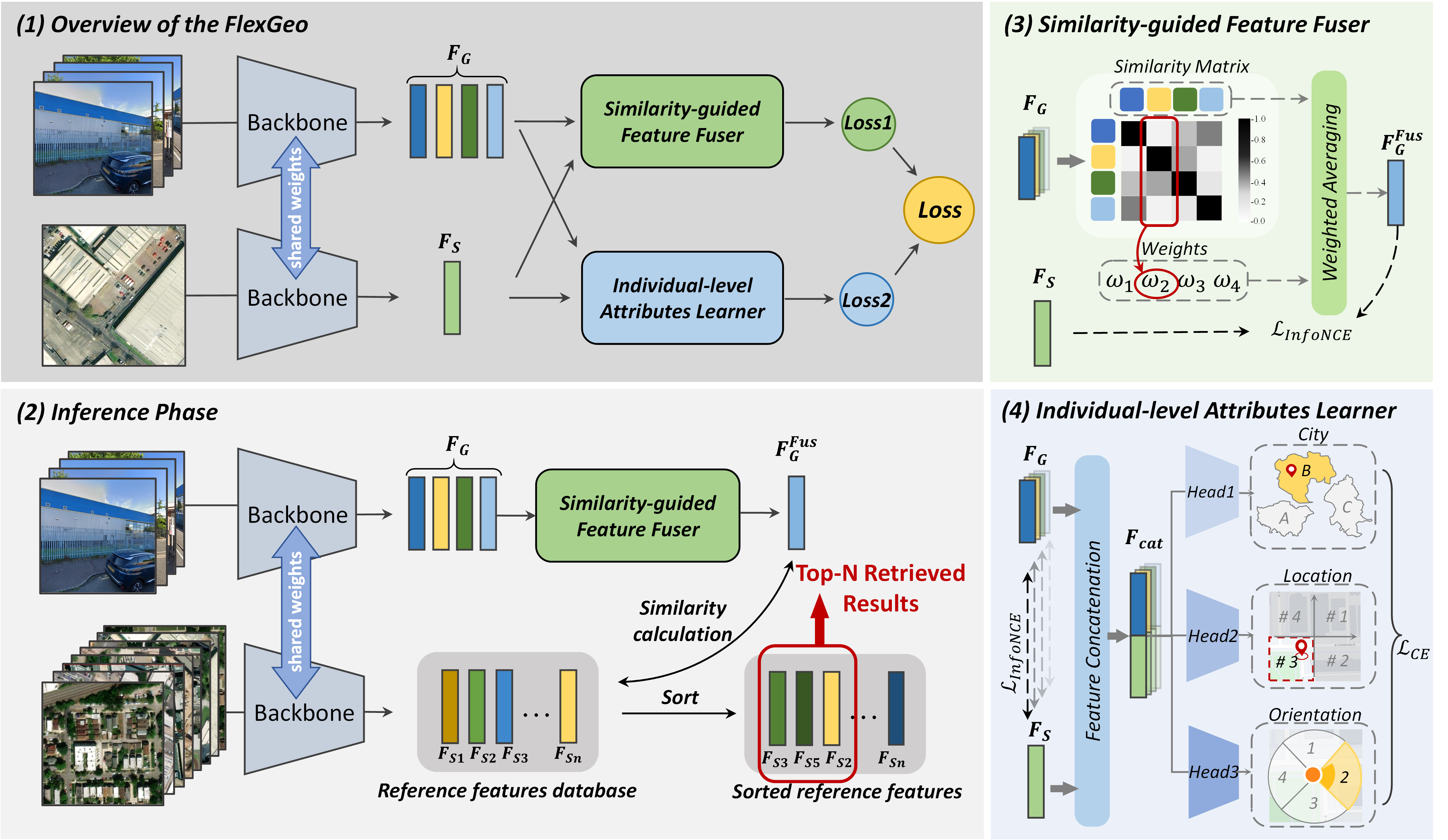}
    \vspace{-0.5em}
    \caption{(1) Overview of the proposed FlexGeo method, featuring two modules: IAL and SFF. (2) Illustration of the FlexGeo inference phase. (3) Detailed illustration of the SFF module. (4) Detailed illustration of the IAL module. }
    \label{fig_FlexGeo}
\vspace{-0.5em}
\end{figure*}

\subsection{FlexGeo Overview}

The complete network structure is illustrated in \cref{fig_FlexGeo}. Our method employs a contrastive learning architecture with separate branches for satellite and ground images. In the satellite branch, the backbone extracts satellite features directly for retrieval. In the ground branch, it processes each ground image independently, generating features that then pass through two modules: SFF and IAL. The SFF fuses features based on similarity to create fused ground features with distinct characteristics, while the IAL combines ground and satellite features for multi-task learning, further enhancing scene understanding.

FlexGeo employs ConvNeXt \cite{liu2022convnet} as its backbone for feature extraction across two branches with shared weights. For the ground branch input images, \(I_G \in \mathbb{R}^{B\times N\times H\times W\times 3}\), where \(B\) is the batch size,  \(N\) is the set size, \(H\) and \(W\)are image sizes, and 3 represents RGB channels. ConvNeXt processes each ground image independently, resulting in feature tensors \(F_G \in \mathbb{R}^{B\times N \times C}\), where \(C\) is the feature dimension. Satellite images are processed similarly, producing satellite features \(F_S \in \mathbb{R}^{B\times C}\). These extracted features are then passed into the SFF and IAL modules. 

In SFF, we fuse the ground features within each set into a single feature \(F_{G}^{Fus} \in \mathbb{R}^{B\times C}\), using feature similarity to guide fusion weights and retain characteristics of the ground images. The InfoNCE loss \cite{oord2018representation} is then calculated between these fused features and their corresponding satellite features. In IAL, rather than treating ground images as a unified group, we apply contrastive constraints between each ground image and satellite images individually, using the InfoNCE loss. Each ground feature is concatenated with its satellite feature along the channel dimension, forming \(F_{cat} \in \mathbb{R}^{B\times N \times 2C}\). Inspired by \cite{deuser2023orientation}, we process these concatenated features through classification heads to predict geo-attributes, with a Cross-Entropy loss \cite{shannon1948mathematical} calculated for each prediction.
During inference (see \cref{fig_FlexGeo}-(2)), only the fused ground features \(F_{G}^{Fus} \in \mathbb{R}^{B\times C}\) and the satellite features \(F_S \in \mathbb{R}^{B\times C}\) are used for retrieval, with IAL applied only during training.

\subsection{Similarity-guided Feature Fuser}

To address the challenge of fusing ground features without sequential relationships, we developed SFF (see \cref{fig_FlexGeo}-(3)). In Set-CVGL, we aim for feature fusion beyond a simple aggregation. Each ground image’s semantic and geometric details are encoded into a feature vector. The importance of each feature is assessed by its distinctiveness within the set. Therefore, we design the SFF to dynamically adjust weights based on feature similarity. The guiding principle is that features with high similarity to others in the set are assigned lower weights, while those that stand out or have unique characteristics receive higher weights. For instance, if a set of images mainly depicts buildings and vegetation, but one captures a river, the feature representing the river should receive greater weight during fusion, as it uniquely signals the presence of a river in the scene.

For each set \((I)\) with ground features  \(F_G^{(I)} \in \mathbb{R}^{N \times C}\), SFF first calculates the cosine similarity between feature vectors, yielding a similarity matrix \(Sim^{(I)} \in \mathbb{R}^{N\times N}\). This matrix is then scaled and normalized to create the adjacency matrix  \(\tilde{A}^{(I)} \in \mathbb{R}^{N\times N} \), with each entry indicating the connection strength between features. Using \(\tilde{A}^{(I)} \), SFF assigns weights based on feature distinctiveness. The sum of each feature's connection strengths in \(\tilde{A}^{(I)} \) yields \(A_{sum}^{(I)} \in \mathbb{R}^{N}\), where each \(A_{sum,i}^{(I)}\) reflects the overall similarity of feature \(i\) to others. The final weight for each feature is computed as \cref{eq:weight}:
\begin{equation}
W_i^{(I)} = {\frac{1}{\left ({A_{sum,i}^{(I)}}\right )^{scale}} } 
\label{eq:weight}
\end{equation}
Here, \(scale\) is a hyperparameter adjusting similarity influence. Features with higher similarity scores receive lower weights to reduce redundancy, while distinctive features receive higher weights, giving them more prominence in the final fused feature. Fused features are then obtained via weighted averaging, as in \cref{eq:fusion}:
\begin{equation}
F_{G}^{Fus(I)} = \sum_{i=1}^{N}\left ( F_{G,i}^{(I)}\odot \frac{W_i^{(I)}}{\sum_{j=1}^{N}W_{j}^{(I)} } \right )
\label{eq:fusion}
\end{equation}
SFF performs this fusion for each set, producing fused ground features \(F_{G}^{Fus} \in \mathbb{R}^{B\times C}\) that integrate diverse information while highlighting unique attributes. This fusion approach operates without sequence dependence and adapts to varied input sizes.

\subsection{Individual-level Attributes Learner}

While SFF preserves diverse information from multiple queries, some information loss still exists during fusion. To address this, we introduce IAL, which optimizes the backbone by leveraging geo-attributes between ground and satellite images. It ensures full utilization of each ground feature in training, even if its contribution to the final fused feature is minor.

As shown in \cref{fig_FlexGeo}-(4), IAL introduces additional optimization objectives to enhance feature utilization: 
First, in addition to the contrast between fused ground features and satellite features, IAL constructs extra contrastive constraints by computing the InfoNCE loss between each individual ground image and the satellite image. This ensures that each ground image is treated independently within the contrastive learning framework. Second, IAL incorporates prior geographic knowledge to deepen model understanding. Each ground feature is concatenated with its satellite feature along the channel dimension, forming concatenated features \(F_{cat} \in \mathbb{R}^{B\times N \times 2C}\). These are then processed by three classification heads to predict their geo-attributes, including city location, relative position, and orientation.

For city location prediction, the model learns the city of each image, helping it distinguish visual traits across cities. For location prediction, we use a two-dimensional coordinate system centered on each satellite image, with quadrants representing distinct location categories for the model to classify each ground image. Orientation prediction assigns each ground image in one of four classes, based on the angle between the camera’s capture direction and the upward direction of the satellite image. Learning these spatial and directional cues helps the model accurately match between cross-view images.

The loss function for the IAL is expressed as follows:
\begin{equation}
\mathcal{L}_{IAL} =  \mathcal{L}_{single} + \lambda_1 \cdot \mathcal{L}_{city} + \lambda_2 \cdot \mathcal{L}_{pos} +\lambda_3 \cdot \mathcal{L}_{ori}
\label{eq:IAL_loss}
\end{equation}
where \(\mathcal{L}_{single}\) represents the InfoNCE loss between each individual ground image and the satellite image. \(\mathcal{L}_{city}\), \(\mathcal{L}_{pos}\) and \(\mathcal{L}_{ori}\) represent the Cross-Entropy loss for prediction of city, position, and orientation, respectively. \(\lambda_1\), \(\lambda_2\), and \(\lambda_3\) are hyperparameters for balancing the losses.

By incorporating these contrastive constraints and geographical predictions, IAL mitigates potential information loss from feature fusion. Even if a ground image has minimal impact on the fused feature, it is still critical for optimizing the model, ultimately enhancing the localization capability.

\subsection{Loss Function}
During training, the overall loss function is denoted as \(\mathcal{L}oss\), as defined below:
\begin{equation}
\mathcal{L}oss = \mathcal{L}_{set} + \lambda \cdot\mathcal{L}_{IAL} 
\label{eq:loss}
\end{equation}
where \(\mathcal{L}_{set}\) represents the InfoNCE loss between fused ground features and satellite features. \(\lambda\) is hyperparameter used to balance the losses.

\section{Experiments}

\subsection{Datasets and Experimental Settings}
We evaluated FlexGeo on our SetVL-480K and two public Sequence-CVGL datasets: SeqGeo \cite{zhang2023cross} and KITTI-CVL \cite{shi2022cvlnet}. For SetVL-480K and SeqGeo, we use the Recall@K-N and AP-N as evaluation metrics. For KITTI-CVL, we adhere to its protocols, using Recall@K-N within 10 meters to measure the retrieval success rate.

The model uses ConvNeXt-Base pre-trained on ImageNet as the backbone, optimized with AdamW over 80 epochs. The hyperparameter \(scale\) in \cref{eq:weight} is set to 2, while \(\lambda_1\), \(\lambda_2\), and \(\lambda_3\) in \cref{eq:IAL_loss} are set to 0.1, 0.2, and 0.2, respectively. The hyperparameter \(\lambda\) in \cref{eq:loss} is set to 1.0. All experiments are conducted on 8 NVIDIA V100-32G GPUs. The batch size is set to 64 (query sets paired with reference images), with 4 query images per set during training.

\subsection{Results on SetVL-480K Dataset}

\begin{table*}[!t]
    \centering
    \fontsize{7pt}{9pt}\selectfont 
    \resizebox{\textwidth}{!}{
    \begin{tabular}{l|c|c|ccccc}
    \toprule
          Method  &Query Setting&Publication&  R@1-N&  R@5-N&  R@10-N &R@1\%-N& AP-N\\  \midrule
 TransGeo~\cite{zhu2022transgeo}&\multirow{4}{*}{\makecell{Single Image \\ (N = 1)}}& CVPR'2022& 1.34& 5.26& 9.12& 51.56&-\\
 FRGeo~\cite{zhang2024aligning}& & AAAI'2024& 1.75& 7.38& 12.58& 60.95&-\\
          Sample4Geo~\cite{deuser2023sample4geo}  &&ICCV'2023&  16.86&  38.95&  47.71&85.40& 22.24\\
 FlexGeo(Ours)& & -& \underline{18.05}& \underline{41.57}& \underline{52.87}& 82.83&\underline{23.67}\\
         \midrule
  SeqGeo~\cite{zhang2023cross}  &\multirow{3}{*}{\makecell{Multiple Images \\ (N = 4)}}&WACV'2023& 3.43& 12.82& 20.94&80.57&6.46\\
          GAMa-Net~\cite{vyas2022gama}  &&ECCV'2022&  5.79&  19.94&  31.53&\underline{91.07}& 9.95\\
          FlexGeo(Ours)  &&-&  \textbf{39.48}&  \textbf{70.23}&  \textbf{80.04} &\textbf{95.85}& \textbf{46.36}\\
    \bottomrule
    \end{tabular}}
     \caption{Comparison on SetVL-480K. N=1 indicates using single images as queries, while N=4 indicates using image sets of size 4 as queries. The best results are highlighted in \textbf{bold}, and the second-best results are \underline{underlined}.}
     \label{tab:results on MVL}
    \vspace{-0.5em}
\end{table*}

We compared FlexGeo with several state-of-the-art methods from both S-CVGL and Sequence-CVGL. For S-CVGL (TransGeo~\cite{zhu2022transgeo}, FRGeo~\cite{zhang2024aligning}, and Sample4Geo~\cite{deuser2023sample4geo}), one-to-one image correspondences are used in training, and a single image from each query set is selected randomly for testing. For Sequence-CVGL methods (SeqGeo~\cite{zhang2023cross} and GAMa-Net~\cite{vyas2022gama}), query images are arranged in random order during both training and testing, as SetVL-480K's images do not follow a sequence. Our FlexGeo is trained on query image sets and evaluated under two conditions: with single images as queries and with image sets.

As shown in \cref{tab:results on MVL}, FlexGeo achieves best results across all evaluation metrics when using multiple images as queries, with a Recall@1-N of 39.48\%. This demonstrates FlexGeo’s effectiveness in handling unordered image sets, outperforming other methods. When tested with single image queries, FlexGeo remains competitive. Despite not using the SFF module in testing, the results indicate that IAL’s multi-task learning strategy still enhances localization accuracy of S-CVGL. 

Although GAMa-Net and SeqGeo use multiple images as queries, their performance still falls behind Sample4Geo and FlexGeo with single images. This can be attributed to: i) the contrastive learning framework used by Sample4Geo and FlexGeo, which enhances discriminative representation; and ii) the reliance of SeqGeo and GAMa-Net on sequential relationships, leading to a substantial performance drop with unordered queries. These results further highlight the limitations of existing Sequence-CVGL methods in addressing the Set-CVGL task and underscore FlexGeo’s robustness and adaptability.

\subsection{Results on Sequence-CVGL Datasets}

\begin{table}[!t]

    \centering
    \resizebox{\linewidth}{!}{
    \begin{tabular}{l|c|cccc}
    \toprule
         Method &Publication&  R@1-N&  R@5-N&  R@10-N& R@1\%-N\\
         \midrule
         SeqGeo~\cite{zhang2023cross}&WACV'2023&  1.80&  6.45 &  10.36 & 34.38 \\
 GAReT~\cite{pillai2024garet}&ECCV'2024& 3.34& 11.19& 17.18&44.39\\
         FlexGeo(Ours) &-&  \textbf{3.51} &  \textbf{14.22} &  \textbf{20.77} & \textbf{48.53} \\
    \bottomrule
    \end{tabular}
}
 \caption{Comparison on the SeqGeo Dataset. In SeqGeo, N=7.
 \label{tab:results on SeqGeo}}
\end{table}

\begin{table}[!t]
    \centering
    \resizebox{\linewidth}{!}{
    \begin{tabular}{l|c|cccc}
    \toprule
         Method &Test set&  R@1-N&  R@5-N&  R@10-N& R@100-N\\
         \midrule
         CVLNet~\cite{shi2022cvlnet}&\multirow{2}{*}{Test-1}&  21.80&  \textbf{47.92}&  \textbf{64.94}& \textbf{99.07}\\
 FlexGeo$^{\dagger}$(Ours)&& \textbf{41.00}& 44.32& 51.48&84.43\\
 \midrule
 CVLNet~\cite{shi2022cvlnet}&\multirow{2}{*}{Test-2}& 12.90& 27.34& 38.62&\textbf{85.00}\\
         FlexGeo$^{\dagger}$(Ours)&&  \textbf{19.61}&  \textbf{32.15}&  \textbf{39.87}& 81.08\\
    \bottomrule
    \end{tabular}}
     \caption{Comparison on KITTI-CVL Dataset. In KITTI-CVL, N=4. $^{\dagger}$indicates that in KITTI-CVL the location label is decided by the first image in each sequence, while FlexGeo does not use this information.
 \label{tab:results on KITTI-CVL}}
\end{table}

As shown in \cref{tab:results on SeqGeo,tab:results on KITTI-CVL}, FlexGeo is evaluated on two public Sequence-CVGL datasets and compared with each dataset’s state-of-the-art methods. While these methods utilize sequential relationships among query images, FlexGeo processes them as unordered sets, achieving competitive results without relying on prior conditions.

\noindent\textbf{SeqGeo Dataset.} The SeqGeo dataset presents challenges due to its composition: approximately 70\% of the images are from suburban areas, and the remaining 30\% are from urban environments. This heavy presence of suburban imagery, lacking distinct geographical cues, makes localization difficult. Despite this, FlexGeo outperforms other methods (see \cref{tab:results on SeqGeo}), achieving a Recall@1-N of 3.51\% over SeqGeo (1.90\%) and GAReT (3.34\%). 

\noindent\textbf{KITTI-CVL Dataset.} In KITTI-CVL, each sequence is labeled by the location of its first image, while FlexGeo operates without knowing which image appears first. Considering the short sequence lengths in KITTI-CVL, we continue to use the position labels provided with the dataset. As shown in \cref{tab:results on KITTI-CVL}, FlexGeo achieves higher Recall@1-N than CVLNet across both test sets (including the simpler Test-1 and the more challenging Test-2 with interference images). However, for Recall@5-N, Recall@10-N, and Recall@100-N, the advantage of FlexGeo lessens and is sometimes surpassed. We attribute these results to different localization objectives: KITTI-CVL focuses on matching the first image in a sequence, while FlexGeo evaluates the sequence as a whole. If the correct reference image aligns well with all query images, FlexGeo assigns it a high score. Conversely, if the reference image matches only the first query image but does not align well with the others, FlexGeo assigns it a low score, possibly excluding it from the top-5, top-10, or even top-100 results. This stricter criterion prioritizes references that consistently match across all query images, enabling FlexGeo to excel in Recall@1 but reducing its tolerance for alternative matches. More detailed analysis is included in the supplementary materials.

Overall, these results show FlexGeo’s robust localization capabilities across varied scenarios, supporting its potential for real-world applications.

\begin{figure}[t]
    \centering
    \includegraphics[width=1\linewidth]{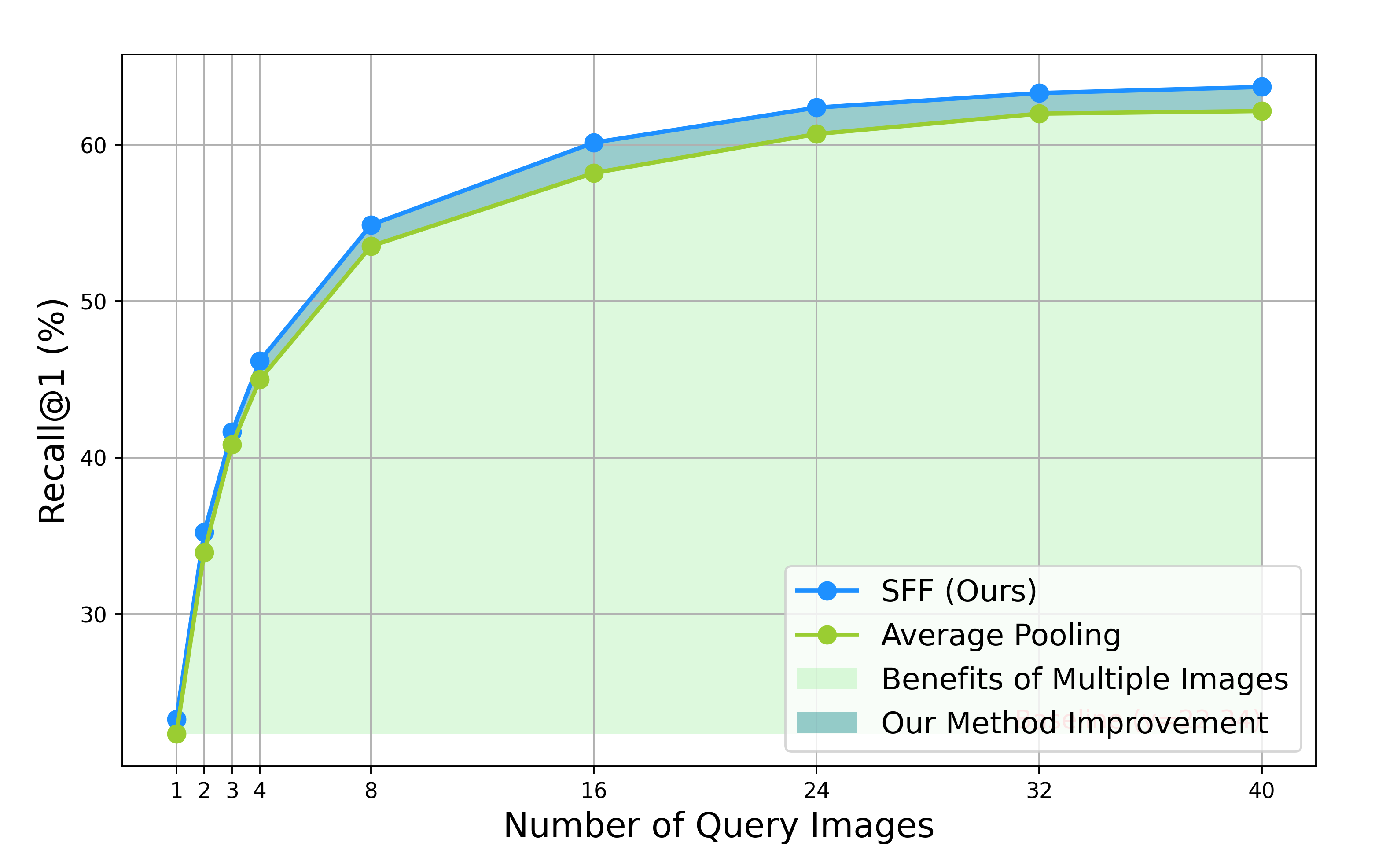}
    \caption{Evaluation of the number of query images. The blue line represents localization results using the SFF module for feature fusion, while the green line shows results with average pooling. }
    \label{fig_accuracy_improvement}
\vspace{-0.5em}
\end{figure}

\begin{table}[!t]
    \centering
    \resizebox{\linewidth}{!}{
    \begin{tabular}{l|cccc}
    \toprule
         Method&  R@1-N&  R@5-N&  R@10-N & AP-N\\
         \midrule
         w/o SFF, IAL&  32.31&  63.41&  75.40 & 39.42\\
 w/ SFF& 34.73& 65.75 & 76.95  &41.76\\
 w/ IAL& 38.54 & 69.47 & 79.64  &45.47\\
         \textbf{w/ SFF, IAL (Ours)}&  \textbf{39.48} &  \textbf{70.23} &  \textbf{80.04}  & \textbf{46.36}\\
    \bottomrule
    \end{tabular}
}
 \caption{Ablation study to verify the effect of SFF and IAL. N=4.
 \label{tab:effect of SFF and IAL}}
\end{table}

\subsection{Ablation Studies}

\noindent\textbf{Evaluation of the Number of Query Images.} We evaluated the benefits of using multiple query images for localization. To ensure consistency, we filtered scenes in SetVL-480K containing at least 40 query images and tested localization accuracy with \(N\) (the number of query images) ranging from 1 to 40 (see \cref{fig_accuracy_improvement}). We conducted tests on our FlexGeo method and a baseline where the SFF module was replaced with average pooling. The results show that increasing \(N\) consistently improves localization accuracy for both methods. Notably, accuracy rises sharply by about 30\% from \(N=1\) to \(N=8\), with diminishing returns as \(N\) approaches 40. This trend highlights the advantage of using multiple query images from diverse perspectives, particularly in practical CVGL scenarios. Users can substantially boost localization accuracy by capturing just one or a few additional images, demonstrating the practical value of our approach. Additionally, the method using SFF for feature fusion consistently outperforms the average pooling approach across all values of \(N\), demonstrating the SFF module's capability in adaptive feature weight adjustment.

\noindent\textbf{Effect of the SFF and IAL.} The ablation study in \cref{tab:effect of SFF and IAL} evaluates the impact of SFF and IAL on FlexGeo’s performance. Four setups are compared: backbone with only average pooling, SFF alone, IAL alone, and both SFF and IAL. While the baseline model performs reasonably well, incorporating SFF improves metrics across the board, reflecting enhanced feature fusion. IAL alone offers an even greater boost, showing its effectiveness in utilizing individual image geo-attributes for improved localization. The combined use of SFF and IAL achieves the highest scores, confirming that their integration enhances both feature representation and localization ability.

\begin{figure}[h!]
    \centering
    \includegraphics[width=1\linewidth]{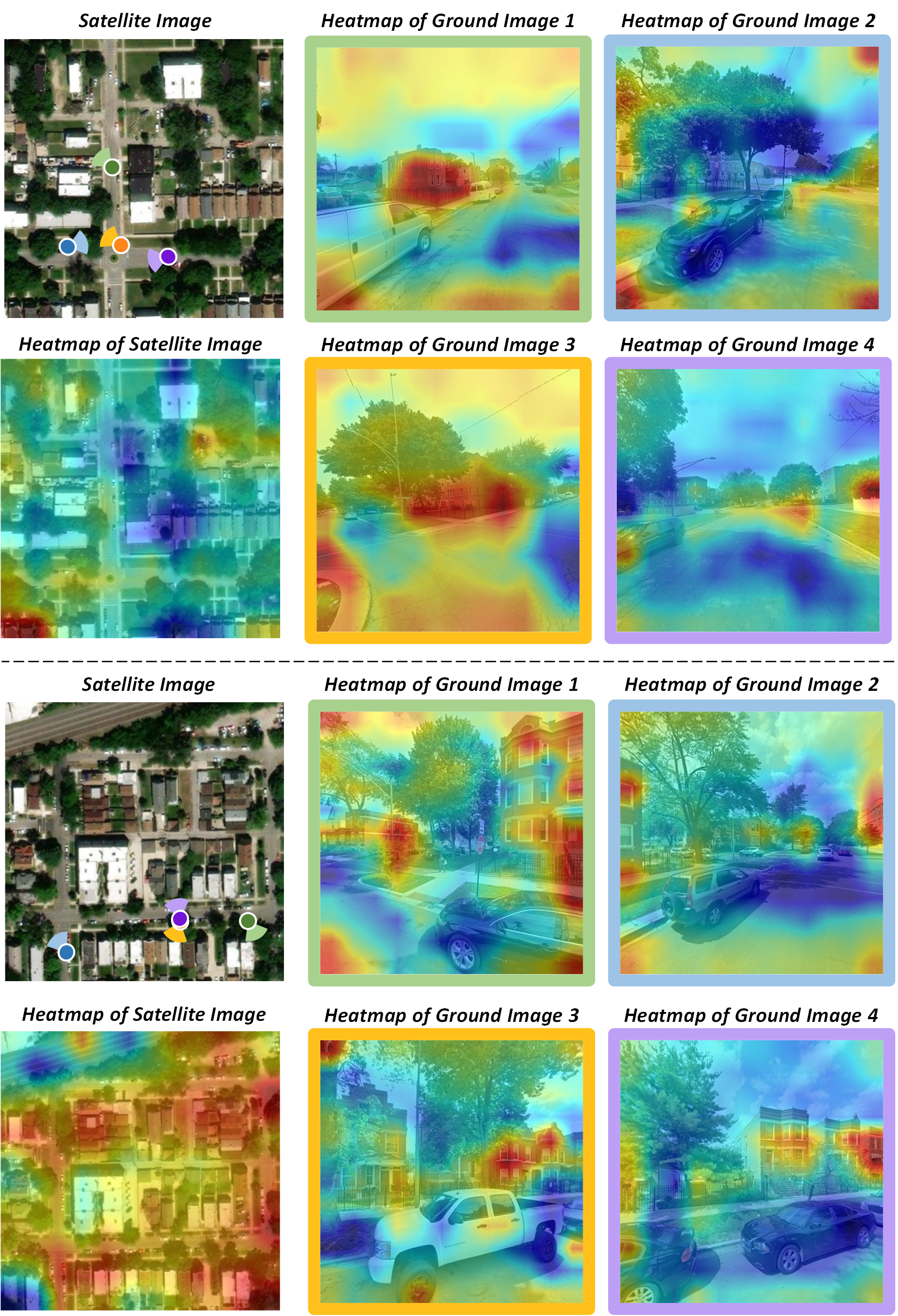}
    \caption{Heatmap visualization of SetVL-480K dataset images generated by the FlexGeo model. The first column contains satellite imagery and its heatmap. The second and third columns are the heatmaps of four corresponding ground images. The sampling positions and orientations of the ground images are marked on the satellite images.}
    \label{fig_heatmaps}
\end{figure}

\noindent\textbf{Visualization.} To qualitatively evaluate FlexGeo, we generate heatmaps of features from the FlexGeo model using test set images in SetVL-480K. As shown in \cref{fig_heatmaps}, we display heatmaps for two satellite images and their corresponding four ground images. For clarity, we mark the sampling positions and orientations of the ground images on the satellite images. The heatmaps demonstrate FlexGeo's ability to focus on distinct, co-visible regions shared across cross-view images, such as buildings, street corners, and vegetation. Notably, the model disregards the transient objects like vehicles, which frequently appear in ground images but do not aid in reliable geo-localization. This selective focus on stable, geographically relevant features underscores FlexGeo’s robustness in cross-view matching, even in complex, dynamic real-world environments.

Additional ablation studies are included in the supplementary materials.

\section{Conclusion}

This study introduces the Set-CVGL task, which improves localization accuracy by using multiple query images from diverse perspectives and locations. To support Set-CVGL, we present SetVL-480K, a benchmark dataset containing 480,000 ground images and corresponding satellite images from six cities across the world. We also propose FlexGeo, a flexible method designed for Set-CVGL that can also adapt to single-image and image-sequence inputs. FlexGeo introduces two key modules: the Similarity-guided Feature Fuser (SFF) and the Individual-level Attributes Learner (IAL). It consistently outperforms existing methods on SetVL-480K and two public Sequence-CVGL datasets: SeqGeo and KITTI-CVL. Furthermore, our experiments demonstrate that using multiple query images with diverse perspectives significantly boosts localization accuracy.


{
    \small
    \bibliographystyle{ieeenat_fullname}
    \bibliography{main}
}

\clearpage
\setcounter{page}{1}
\maketitlesupplementary
\setcounter{section}{0}
\renewcommand\thesection{\Alph {section}}
\renewcommand\thefigure{\arabic{figure}} 
\renewcommand\thetable{\arabic{table}}

\section{Overview}
This supplementary material provides additional information to support and expand on the main content of our paper, divided into three sections:

\noindent\textbf{Analysis of the Set-CVGL Task:} We explore the advantages and challenges of the Set-CVGL task in detail (\cref{sup:sec:task}).

\noindent\textbf{Details of the SetVL-480K Dataset:} We supplement information on the data distribution, organization, and usage of the SetVL-480K dataset (\cref{sup:sec:dataset}).

\noindent\textbf{Expanded Experiments and Analysis:} We provide extended experiments and analysis with FlexGeo, including:
\begin{itemize}
\item Detailed analysis of the results on the KITTI-CVL dataset (\cref{sup:sec:kitti-cvl}).
\item Verification of the effectiveness of different tasks in the IAL module (\cref{sup:sec:IAL}).
\item Selection of the hyperparameter in the SFF module (\cref{sup:sec:hyperparameter}).
\item Visualization of weight adjustment results in the SFF module (\cref{sup:sec:SFF}).
\item Visualization of retrieval results with FlexGeo (\cref{sup:sec:visualization}).
\end{itemize}

\section{Advantages and Challenges of Set-CVGL}
\label{sup:sec:task}
The Set-CVGL task offers significant advantages over traditional single-image and image-sequence approaches by utilizing multiple ground images from varying perspectives and locations. This enables a comprehensive capture of environmental information, improving localization accuracy by overcoming the limitations, such as occlusions or partial views, which are particularly common in complex urban settings. The richer set of visual cues provided by multiple images not only enhances the precision but also improves the robustness of the localization system, making it more adaptable to challenging conditions such as lighting change or obstructed views. 

Set-CVGL serves as a generalized form of CVGL tasks, as methods designed for Set-CVGL are inherently capable of adapting to single-image or image-sequence CVGL scenarios. Moreover, using multiple query images from diverse perspectives aligns with natural human behavior and does not impose a significant burden on users when collecting data. In many real-world situations, such as navigating city streets, exploring new areas, or conducting drone-based surveillance, users naturally capture multiple images from different angles and locations. Additionally, since the collection of query data is user-driven, controlling the radius of the query image sets is feasible and practical. 

Set-CVGL also faces several challenges. Fusing features from multiple images, each captured from different perspectives and locations and often with varying quality, is inherently complex. The fusion process must integrate diverse visual information without losing key details while also avoiding redundancy from similar content. In cases where query images have sequence relationships, positional cues between objects can assist in feature fusion. However, in the Set-CVGL task, there is no sequence relationship between the query images, and they often lack a high degree of overlap, making it impossible to reconstruct the 3D scene using multi-view geometry. Therefore, creating a unified and representative feature becomes even more challenging. Additionally, Set-CVGL methods must handle variable-length inputs, ensuring the model can adapt to different numbers of query images without sacrificing too much computational efficiency. The challenge lies in designing flexible architectures that can scale with the input size while maintaining a balance between precision, speed, and resource usage.

\section{Additional Details of SetVL-480K}
\label{sup:sec:dataset}

\begin{figure*}[t]
    \centering
    \includegraphics[width=1\linewidth]{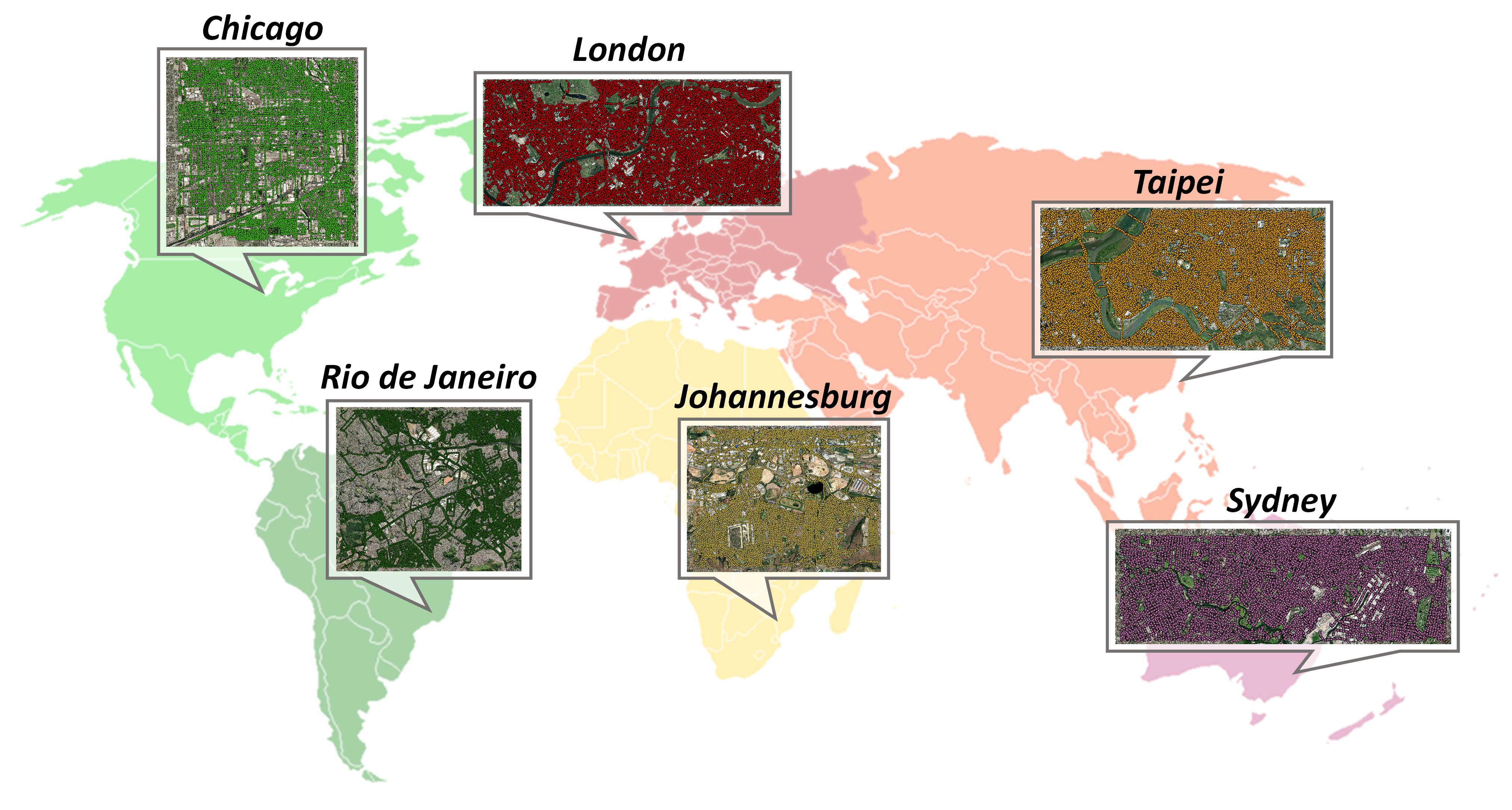}
    \caption{Distribution of images in SetVL-480K. The basic world map is obtained from \url{https://en.wikipedia.org/wiki/Continent}.}
    \label{fig_data_distribution}
\end{figure*}

This section provides a detailed description of the SetVL-480K dataset, including its data distribution, organization, and usage. The dataset consists of data from six cities: Chicago, London, Taipei, Rio de Janeiro, Johannesburg, and Sydney. Each city contains satellite images covering its main urban area as reference images along with 80,000 ground-level images as query images. The coverage area of each city spans approximately 100 \(km^{2}\). To visualize the dataset, we plotted the distribution of ground images overlaid on the corresponding satellite images. As shown in \cref{fig_data_distribution}, the ground images in each city are densely distributed, providing rich query data for geo-localization tasks.

In SetVL-480K, we explicitly record the correspondences between reference and query images. For the training data, users can freely decide how to organize the data. For instance, if a single reference image corresponds to 40 query images, users can choose to train models with 40 one-to-one mappings, 20 one-to-two mappings, or other configurations depending on the method or specific experimental needs.

For the test data, we standardize the set combination of query images for different set sizes. This ensures consistency when evaluating models, as all query sets will contain the same fixed images.

In SetVL-480K, the correspondences between reference and query images are explicitly recorded. For the training data, users have the flexibility to organize the data as needed. For example, if a single reference image corresponds to 40 query images, users can choose to train models using 40 one-to-one mappings, 20 one-to-two mappings, or other configurations tailored to their specific methods or experimental requirements.

For the test data, we standardize the construction of query image sets for different set sizes. This ensures consistent evaluations, as query sets are composed of identical, pre-defined images. 

\begin{table}[!t]

    \centering
    \resizebox{\linewidth}{!}{
    \begin{tabular}{l|ccccc}
    \toprule
         Task&  R@1-N&  R@5-N&  R@10-N& R@1\%-N &AP-N\\
         \midrule
         w/o city, pos, ori&  34.73&  65.75 &  76.95  & 93.53&41.76\\
 w/ city& 35.64& 67.06& 77.78&94.02&42.73\\
         w/ city, pos&  36.43&  67.99&  78.58& 94.20&43.50\\
 w/ city, pos, ori& \textbf{39.48}& \textbf{70.23}& \textbf{80.04} & \textbf{95.85}&\textbf{46.36}\\
     \bottomrule
    \end{tabular}
}
 \caption{ Comparison of different classification tasks with set size \(N = 4\).
 \label{tab:results on 3 predictions}}
\end{table}

\begin{table}[!t]

    \centering
    \resizebox{\linewidth}{!}{
    \begin{tabular}{l|ccccc}
    \toprule
         Scale&  R@1-N&  R@5-N&  R@10-N& R@1\%-N &AP-N\\
         \midrule
         1&  39.34&  70.18&  79.95& 95.67&46.24\\
 2& \textbf{39.48}& \textbf{70.23}& \textbf{80.04} &\textbf{95.85}&\textbf{46.36}\\
         3&  38.74&  69.48&  79.37& 95.59&45.61\\
 4& 37.43& 68.01& 78.10& 95.13&44.27\\
     \bottomrule
    \end{tabular}
}
 \caption{Comparison between the different \(scale\) we select for SFF with set size \(N = 4\).
 \label{tab:results on scale}}
\end{table}

\begin{figure*}[t]
    \centering
    \includegraphics[width=1\linewidth]{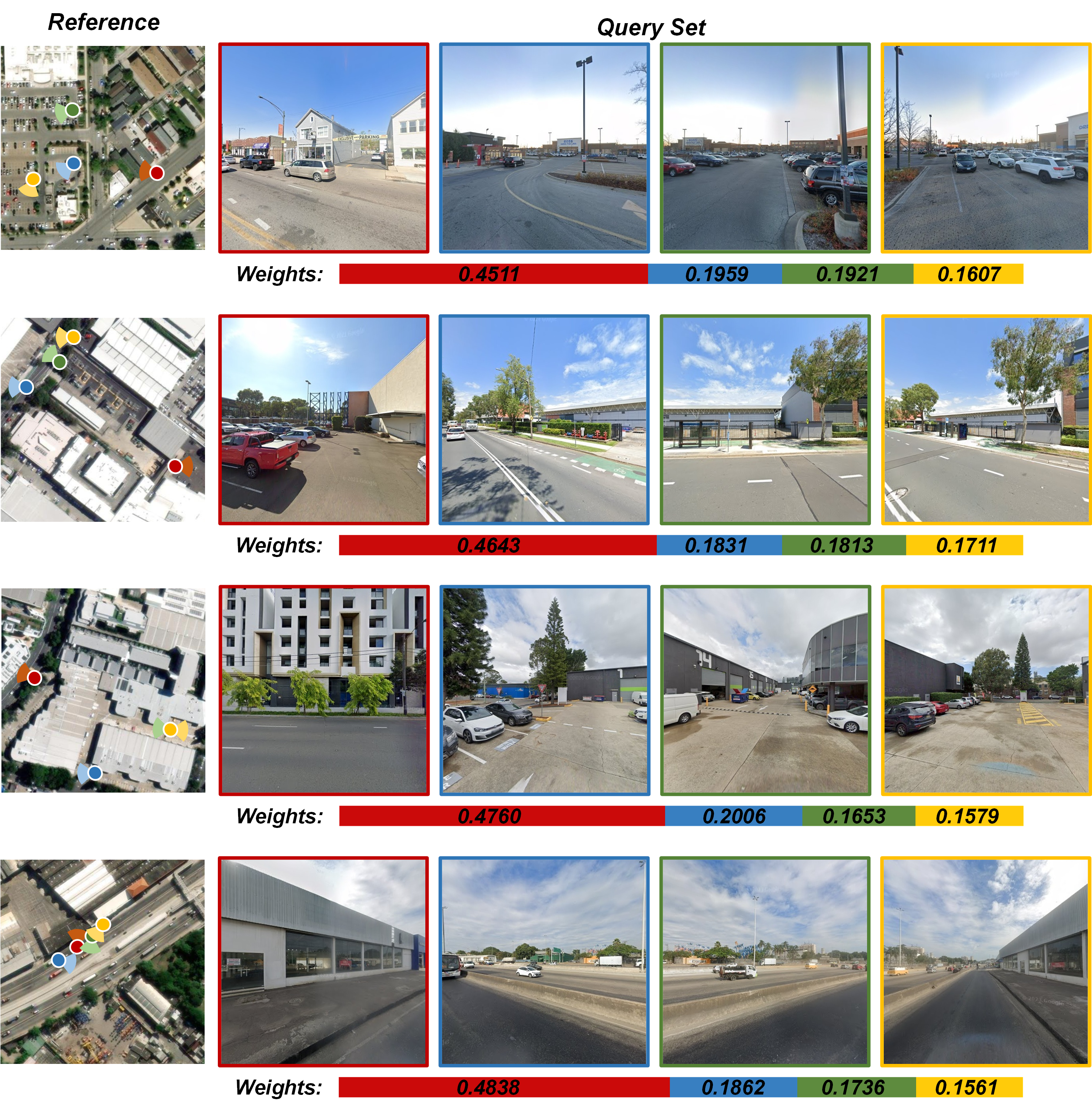}
    \caption{Visualization of weight adjustment results in SFF. The first column shows the reference images, while the second to fifth columns display the corresponding query images. Higher weights are assigned to query images with more distinctive content, indicated by the red square frames and markers. }
    \label{fig_weights}
\end{figure*}

\begin{figure*}[t]
    \centering
    \includegraphics[width=1\linewidth]{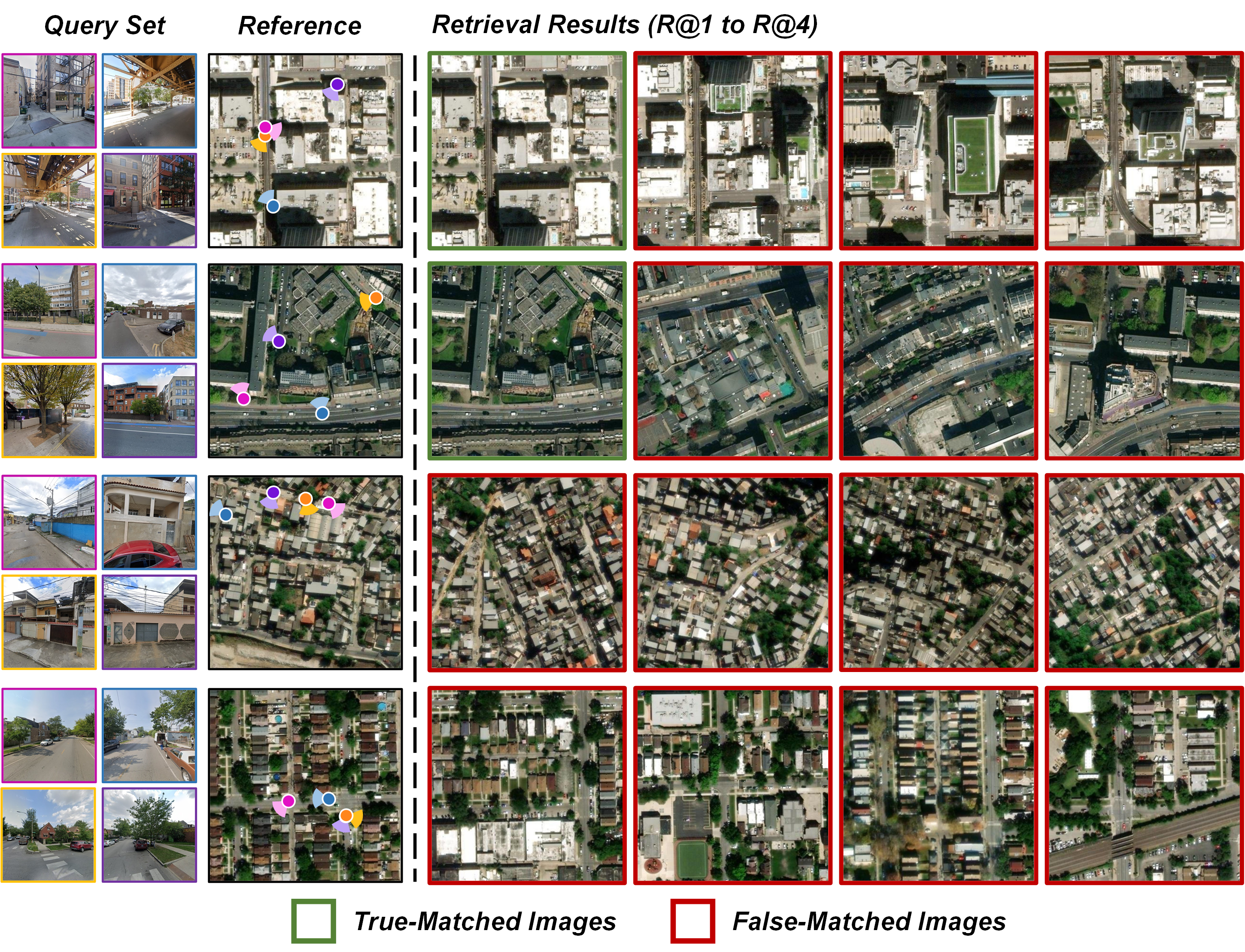}
    \vspace{-1.0em}
    \caption{Visualization of retrieval results. The top two rows depict successful localization cases, where the green boxes indicate correct reference images and red boxes represent incorrect ones. The bottom two rows show failed localization cases. Sampling positions and orientations of the ground images are marked on the satellite images.}
    \label{fig_retrieval_res}
     \vspace{-1.0em}
\end{figure*}

\section{Additional Details of FlexGeo}
\label{sup:sec:additional_flexgeo}

\subsection{Analysis of Results on KITTI-CVL}
\label{sup:sec:kitti-cvl}
In this section, we provide a detailed analysis of results on the KITTI-CVL dataset. Our method shows exceptional Recall@1 performance but is surpassed at higher recall thresholds (e.g., Recall@10, Recall@100). We attribute this to the first-image-focused labeling in KITTI-CVL. Specifically, the dataset is designed to improve the localization accuracy of a single query image \(Q_{1}\) by incorporating additional sequence images \(Q_{2},...,Q_{N}\). In other words, the objective of the dataset is to enhance single-image localization through the inclusion of sequence images. Consequently, the reference image label in this dataset is determined solely by the first query image \(Q_{1}\).

This setting introduces a potential mismatch: the reference image for the sequence may not align well with the additional images in the sequence. For example, consider four query images \(Q_{1},Q_{2},Q_{3},Q_{4}\), whose corresponding reference images are \(R_{1},R_{2},R_{3},R_{4}\). In this dataset, the sequence label is \(R_{1}\), regardless of whether \(R_{2},R_{3},R_{4}\) are consistent with \(R_{1}\). To quantify this inconsistency, we analyzed the test set for \(N = 4\) and found that 33.08\% of image sequences have \(R_{1}\) differing from \(R_{2}\), \(R_{3}\), and \(R_{4}\). This indicates that a significant portion of sequences contain inconsistent reference labels across their query images.

However, FlexGeo treats the query image sequence as a unified whole. It assigns high scores to reference images that align well with all query images in the sequence. Conversely, if the reference matches only the first query image but not so well with others in the sequence, FlexGeo assigns it a low score, potentially excluding it from the top-5, top-10, or even top-100 results. This stricter criterion prioritizes references that consistently match all query images, which explains FlexGeo's superior performance in Recall@1 but its reduced tolerance for alternative matches.

\subsection{Effectiveness of Multi-Task Learning in IAL}
\label{sup:sec:IAL}
To evaluate the impact of multi-task learning in IAL, we perform ablation studies on the three classification tasks: city location prediction, relative position prediction, and orientation prediction. As shown in \cref{tab:results on 3 predictions}, incorporating these tasks improves localization accuracy, indicating that they enhance the model's scene perception. Among these tasks, orientation prediction yields the most significant improvement in localization accuracy. This suggests that for limited-FOV query images in CVGL tasks, understanding the shooting direction is particularly beneficial for accurate localization.

\subsection{Hyperparameter Selection for SFF}
\label{sup:sec:hyperparameter}

To determine the optimal hyperparameter for the Similarity-guided Feature Fuser (SFF), we evaluate the effect of different values of \(scale\) on the SetVL-480K dataset. The hyperparameter \(scale\) controls the influence of similarity in the feature fusion process, directly affecting how features are weighted during fusion. As shown in \cref{tab:results on scale}, our method achieves the best performance with \(scale = 2\). Based on this, we set \(scale = 2\) for all our experiments.

\subsection{Visualization of Weight Adjustment in SFF}
\label{sup:sec:SFF}
To better understand the role of SFF in feature fusion, we present representative examples of weight adjustments. As shown in \cref{fig_weights}, we select cases where the weight distribution among query images is notably uneven (i.e., one query image has a weight greater than 0.4). In these examples, the assigned weight for each query image is displayed, along with its corresponding position and orientation on the satellite reference image.

The results demonstrate that SFF assigns higher weights to query images that are more distinctive compared to others in the set. For instance, images containing unique objects or significantly different content tend to receive greater attention, as indicated by the red highlighted weights in \cref{fig_weights}. This behavior helps SFF avoid information redundancy caused by repeated object types or overlapping content captured across multiple query images, ensuring that the most informative features are emphasized during the fusion process.

\subsection{Visualization of Retrieval Results}
\label{sup:sec:visualization}
As a qualitative evaluation, we present retrieval results of FlexGeo on the SetVL-480K dataset. Specifically, we show two successful localization cases (where the first retrieved image matches the correct reference) and two failed localization cases (where none of the top 50 retrieved images include the correct reference).

As illustrated in \cref{fig_retrieval_res}, FlexGeo demonstrates strong discrimination capability in scenes with simple layouts, such as those dominated by large, distinct objects. In these cases, even when there are visually similar reference images, FlexGeo accurately identifies the correct one. For instance, in the successful localization cases, the top four retrieval results are visually very similar but still include the correct reference.

However, in scenes with complex layouts—such as residential areas featuring small and numerous buildings—FlexGeo faces challenges. While it retrieves images highly similar to the target scene, it struggles to distinguish the exact match among these candidates. This limitation highlights the need for further improvements in discriminating fine-grained differences in complex environments.

\end{document}